\documentclass[runningheads]{llncs}
\usepackage{graphicx}
\usepackage{amsmath,amssymb} 
\usepackage{color}
\usepackage{subfigure}
\usepackage[width=122mm,left=12mm,paperwidth=146mm,height=193mm,top=12mm,paperheight=217mm]{geometry}

\begin{document}
\pagestyle{headings}
\mainmatter
\title{Learning to Rank Binary Codes} 

\titlerunning{Learning to Rank Binary Codes}

\authorrunning{J. Feng, W. Liu, Y. Wang}

\author{Jie Feng\inst{1}, Wei Liu\inst{2}, Yan Wang\inst{3}}    
\institute{Dept. of Computer Science, Columbia University
\and IBM T. J. Watson Research Center
\and Dept. of Electrical Engineering, Columbia University}

\maketitle

\begin{abstract}
Binary codes have been widely used in vision problems as a compact feature representation
to achieve both space and time advantages. Various methods have been proposed to learn data-dependent
hash functions which map a feature vector to a binary code. However, considerable data information
  is inevitably lost during the binarization step which also causes ambiguity in measuring sample similarity
  using Hamming distance. Besides, the learned hash functions cannot be changed after training, which makes
  them incapable of adapting to new data outside the training data set. To address both issues,
  in this paper we propose a flexible bitwise weight learning framework based on the binary codes obtained
  by state-of-the-art hashing methods, and incorporate the learned weights into the weighted Hamming distance computation.
  We then formulate the proposed framework as a ranking problem and leverage the Ranking SVM model
  to offline tackle the weight learning. The framework is further extended to an online mode
  which updates the weights at each time new data comes, thereby making it scalable to large and dynamic data sets.
  Extensive experimental results demonstrate significant performance gains of using binary codes with bitwise weighting
  in image retrieval tasks. It is appealing that the online weight learning leads to comparable
  accuracy with its offline counterpart, which thus makes our approach practical for realistic applications.

\end{abstract}

\section{Introduction}

\begin{figure*}[ht!]
\centering
\includegraphics[width=\linewidth]{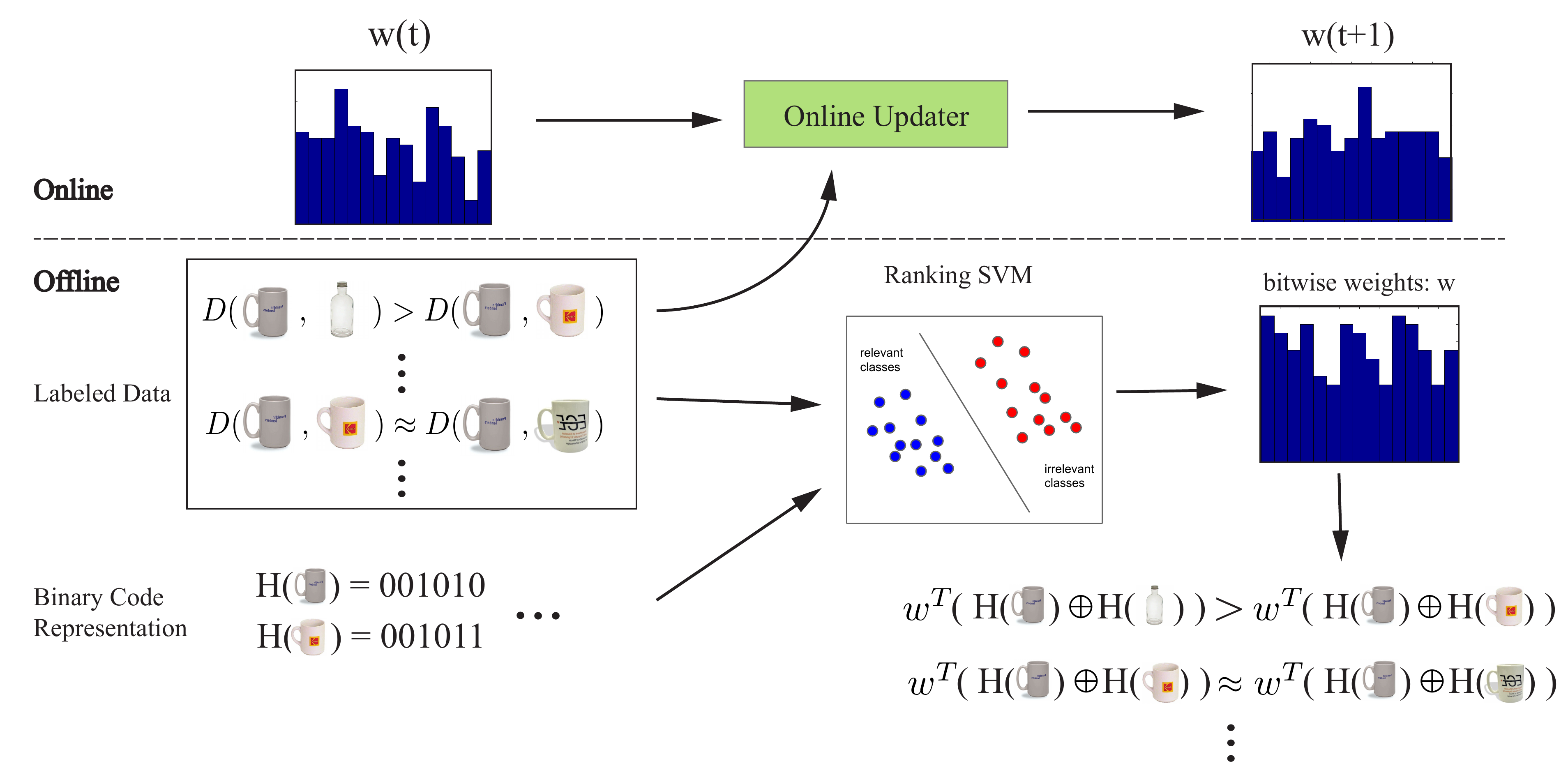}
\caption{Binary code weight learning flowchart in the example of image search.}
\label{flowchart}
\end{figure*}

With easy access to large amount of data nowadays, similarity search requires good balance between storage and efficiency while maintaining good accuracy. To cope with web scale data, binary code representation is receiving more and more interest from academic research and industrial development, particularly for applications like content-based image and video search~\cite{jegou2010improving}~\cite{videohashing}, object recognition~\cite{smallcodes}~\cite{google}, local descriptor matching~\cite{orb}~\cite{boostingkeypoint}, etc. The code itself could be represented as a sequence of binary bits, combined with Hamming distance measure, bringing both time and storage efficiency to Approximate Nearest Neighbor (ANN) search even in high dimensions. This makes binary code representation a practical approach to tackle previously intractable problems.

Due to the recent popularity of learning-based hashing, compact yet informative binary code could be obtained via unsupervised or supervised ways. Although binary codes have drawn significant attention, two major concerns exist in this representation. The first one is the distance ambiguity of binary codes while using Hamming distance. Due to the binarization process to obtain the code, a considerable amount of information is lost. The measurement from Hamming distance can only take discrete values up to the number of code length. It's not uncommon to have a few neighbors share the same distance from a query. To get a finer ranking of these candidates, a stand-alone re-ranking method is usually applied, e.g. RANSAC in image search~\cite{ransac}, reducing the efficiency and increasing the complexity of the search algorithm. Second, in real applications, it is very often for existing data sets to expand with new data coming, which may result in different feature distributions as seen before, and thus affect the efficacy of pre-trained hash functions. However, to update the hash functions, existing methods usually require a complete retraining process on the new data set and recomputing the binary codes for all the samples. This dramatically reduces the flexibility of hashing based search algorithms on frequently updated data sets.

Our contribution in this paper is to propose a unified binary code weighting framework to tackle the above problems. By learning weights for each code bit, we transform the standard Hamming distance to a real-valued weighted Hamming distance, which could serve as a re-ranking mechanism for candidates filtered by the standard Hamming distance, at a minor computational cost. We use human labeled rank pairs as training data. The weight learning problem is cast into a learning-to-rank formulation and the sophisticated Ranking SVM~\cite{rankingsvm} model is utilized. The model is trained such that binary codes of samples from the same class are expected to have smaller weighted Hamming distance than those from different classes. To ensure a valid distance measure for ranking, a non-negative constraint is imposed to the learned weights. Furthermore, we extend the weighting algorithm to an online version using Passive-Aggressive algorithm~\cite{crammer2006online} to allow efficient incremental weight updates. Fig.\ref{flowchart} illustrates the process of weight learning in both offline and online modes. This framework is general enough to be applied to different given binary code base.\footnote{Binary code representation, binary code base are used interchangeably in this paper to denote the binary code directly produced from hash functions.} Experiments are conducted on large real world image data sets and the results demonstrate the effectiveness of the weighting method in producing significant performance gain beyond original binary codes. We also find that the online learning mode is able to achieve comparable performance with its offline counterpart.

\section{Related Work}

\textbf{Binary Code Generation:} Learning methods are used to find good hash functions for generating binary codes given a data set. There are two major branches of hashing algorithms in the literature: unsupervised hashing and supervised hashing. Unsupervised hashing relies on the similarity measure in the original feature space. Locality Sensitive Hashing(LSH)~\cite{lsh} simply uses random projections and a thresholding operation to get the binary code. Other works tried to improve LSH by leveraging a given data set. Spectral Hashing(SH)~\cite{sh} finds the projection by analyzing spectral graph partition. 
Iterative Quantization(ITQ)~\cite{itq} further proposed to use orthogonal projections and a rotational variant to minimize quantization error in Hamming space. Multidimensional Spectral Hashing(MDSH)~\cite{mdsh12} reconstructs the affinity between datapoints rather than distances and is guaranteed to reproduce the affinities with increasing number of bits. Isotropic Hashing(ISOH)~\cite{isoh} learns projection functions which produce equal variances for different dimensions. These methods lack the ability to integrate supervision into the code generation process thus maintain limited semantics in the produced binary codes. Supervised Hashing makes use of available labels to learn semantic-aware hash functions, leading to better quality binary codes regarding search accuracy. Kernel-based Supervised Hashing(KSH)~\cite{ksh} connected code inner products with Hamming distance and used a greedy optimization algorithm to find discriminative hash functions. Other supervised hashing methods~\cite{losshash}~\cite{binaryrecons} also show promising results by embedding class information. However, these methods are usually time consuming to learn on large scale data sets.

\textbf{Learning to Rank:} Ranking algorithms aim to find a proper ranking function given some form of training data. Pairwise rank is commonly used to represent knowledge of relative ranks between two samples. Ranking SVM has been widely used in learning to rank and was first introduced in~\cite{rankingsvm} to cast ranking problem into SVM formulation simply by changing the original samples into sample difference. It benefits from the large margin property of SVMs and add semantics into the ranking pairs so that the results could match better with human's expectation. This model has also been applied to many other problems, such as relative attribute~\cite{relataive} and codebook weighting for image search~\cite{codebookweights}.

\textbf{Binary Code Ranking:} Given the limitation of Hamming distance metric, some works have tried to improve it beyond raw binary code by computing bitwise weights. Jiang et al.~\cite{jiang2013query} proposed a query-adaptive Hamming distance by assigning dynamic class dependent weights to hash bits.
Jun et al.~\cite{iccv13rank} leverages listwise supervision to directly learn hash functions to generate binary codes which encode ranking information. However, this approach is non-convex and is sensitive to initialization. WhRank~\cite{weightedhamming} combines data-adaptive and query-adaptive weights in a weighted Hamming distance measure by exploiting statistical properties between similar and dissimilar codes before applying hash functions. It is general to use for different types of binary codes and showed improvement beyond base code using standard Hamming distance. This is the most similar work to ours in the sense of computing a weight vector for each bit. However, the weight learning method used in WhRank lacks a specific optimization goal and are largely based on observations. Most of the above mentioned weighting methods are offline learned and keep static afterwards.


\section{Approach}

We first introduce notations used in the algorithm and then describe the learning process for our problem. Given an image data set $I=\{\langle x_i, c_i \rangle\}^N_{i=1}$, in which $x_i$ is a D-dimension feature vector of the $i$th image and $c_i$ is the corresponding class label. A set of hash functions $H = \{h_k\}^K_{k=1}$ are applied to each feature vector $x_i$ to compress it to a binary code $H_i=(h_1(x_i), ..., h_K(x_i))$, where $h_k(x_i)$ is either $0$ or $1$ and could be denoted as $H_i^k$. The Hamming distance between two codes is computed as $D_H(H_i, H_j))=|\{k|h_k(x_i) \neq h_k(x_j), k=1...K\}|$. Since the code is binary, the bitwise XOR result is identical to the absolute difference between the two code vectors, $|H(x_i)-H(x_j)|$, we call this vector Absolute Code Difference Vector (ACDV), represented as $Acdv(H_i, H_j)=(|h_1(x_i)-h_1(x_j)|, ..., |h_K(x_i)-h_K(x_j)|)$. Our goal is to learn a weight vector $w$ for binary code to reveal the relative importance of each code bit in distance measure. The code distance is then transformed into a weighted Hamming distance: $D_H^w(H_i, H_j)=w^T Acdv(H_i, H_j)$. This weighted Hamming distance produces a real-valued distance which enables direct ranking of all codes in the data set given a query code.

\subsection{Binary Code Weighting}

\begin{figure}[t]
\subfigure[]{
\includegraphics[width=0.3\textwidth, trim=0 0 5 0]{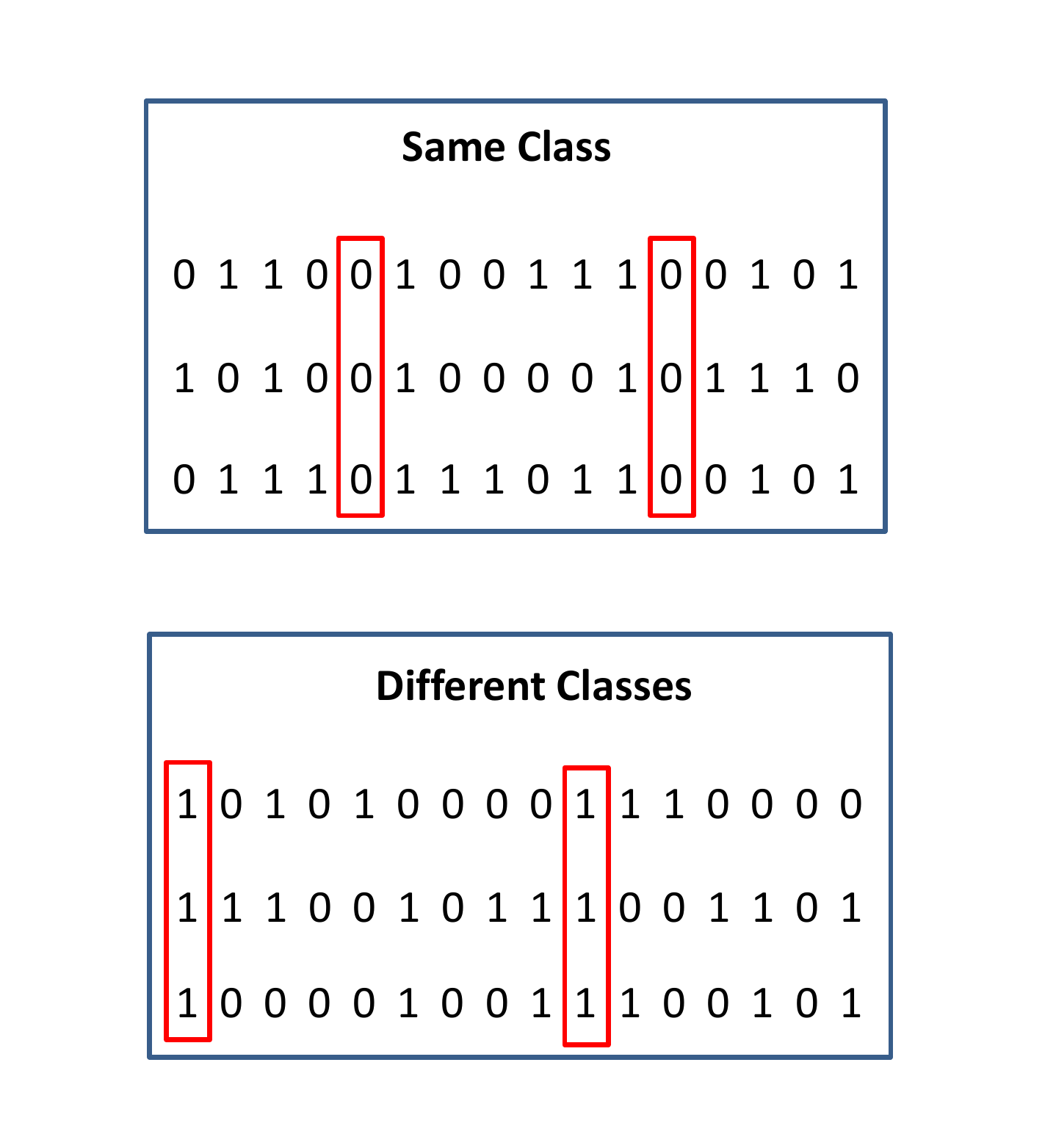}
\label{acdv_sketch}
}
\subfigure[]{
\includegraphics[width=0.3\textwidth, trim=0 0 5 0]{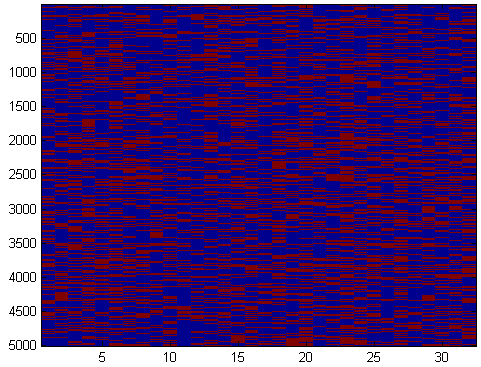}
\label{itq_simcls}
}
\subfigure[]{
\includegraphics[width=0.3\textwidth, trim=0 0 5 0]{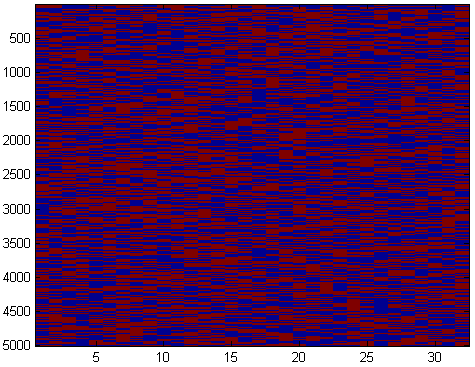}
\label{itq_diffcls}
}
\caption{(Best viewed in color) (a) Conceptual ACDVs between similar and dissimilar code pairs; (b) ACDVs of 32 bits ITQ from the same class; (c) ACDVs of 32bits ITQ codes from different classes. In (b) and (c), each row is an ACDV where red step denotes 1 and blue step denotes 0.}
\label{acdv}
\end{figure}

We analyze the code bit distribution of the ACDVs between any two binary codes.
Assume there are a set of binary code pairs from the same class and another set from different classes. The intuition behind binary code weight learning is illustrated in Fig.\ref{acdv_sketch}.
For codes from the same class, there might be certain bits showing high probability to have the same bit value, resulting corresponding bit in ACDV as 0 (red boxes in the upper part of Fig.\ref{acdv_sketch}); similarly, as for codes from different classes, some bits are more likely to be different, resulting the ACDV bit as 1 (red boxes in the lower part in Fig.\ref{acdv_sketch}). These bits are thus more discriminative to either group semantically similar codes or distinguish dissimilar ones, which should be given higher weights.

We visualize ACDVs for 32 bits ITQ code from MNIST hand-written character data set~\cite{minst}. ACDVs from the same class are shown in Fig.\ref{itq_simcls} and those from different classes are shown in Fig.\ref{itq_diffcls}. It is obvious to see that some columns (bit locations) of ACDVs in Fig.\ref{itq_simcls} are much sparser than other columns indicating these bits are more useful to recognize codes from the same class, thus the weights for these bits should be higher to produce larger distance if two codes differ in these locations. As for Fig.\ref{itq_diffcls}, those columns with more 1's (red) are better to indicate codes from different classes.

A proof-of-concept experiment is present to show whether it is a valid assumption that bitwise weighting could be applied to encode supervision and improve the semantics of a fixed binary code base. The main purpose of the experiment is to learn a weight vector for binary code bits so that codes of the same class could be well separated from codes of different classes. We use LSH~\cite{lsh}, SH~\cite{sh} and ITQ~\cite{itq} to create 64 bits binary codes on a subset of the MNIST. Each digit class in the subset has 1000 grayscale images, among which 70\% is used to for training and 30\% is used for testing. Raw pixel values serve as raw feature for generating binary code. The positive samples are the ACDVs between code pairs from the same class and the negative ones are ACDVs from different classes. A linear SVM is trained to classify these samples. Its prediction accuracy on ACDVs from testing data is shown below in Table.\ref{tab.acdvs_clf}:

\begin{table}[h]
\center{
\begin{tabular}{|c|c|c|c|}
\hline
 Code Type & LSH & SH & ITQ \\
 \hline
 Accuracy & 71.3\% & 83.5\% & 85.7\% \\
 \hline
\end{tabular}
}
\vspace{1em} \caption{Accuracy of ACDVs classification using linear SVM on a subset of MNIST.}
\label{tab.acdvs_clf}
\end{table}

This classification results indicate strongly that we could learn a proper weight vector to improve the discriminative power of binary codes for finer distance measure and ranking.

\subsection{Weight Learning as a Ranking Problem}

We now formulate the weight learning problem in a learning-to-rank paradigm. In order to get semantically discriminative distance measure, we aim to find the weights so that distance between samples from the same class is small and distance between samples from different classes is big. In our case, the weighted Hamming distance is used as distance measure. Given a data set containing binary code and class label pair for each sample, $\chi=\{\langle H(x_i), c_i \rangle\}^N_{i=1}$, we optimize the weights $w$ so to satisfy as many constraints as below:
\begin{equation}\label{constraint1}
\begin{aligned}[l]
    & \forall(i,j,k)\; c_i \neq c_j \: \wedge \: c_i = c_k \\
    & D_H^w(H_i, H_j) > D_H^w(H_i, H_k)
\end{aligned}
\end{equation}

\begin{equation}\label{constraint2}
\begin{aligned}
    & \forall(i,j,k)\; c_i = c_j = c_k \\
    & D_H^w(H_i, H_j) = D_H^w(H_i, H_k)
\end{aligned}
\end{equation}

We adopt the Ranking SVM formulation similar as~\cite{largescaleranking}. Here the ranking is induced to reflect both inter-class separation and intra-class proximity. We will show how the training data is generated in Sec~\ref{train_data}. The complete optimization problem is as follows:
\begin{equation}\label{opt1}
\begin{aligned}
    \min \frac{1}{2}\|w\|_2^2 + C_{\xi} \sum \xi_{ijk}^2 + C_{\gamma} \sum \gamma_{{i'}{j'}{k'}}^2
\end{aligned}
\end{equation}
\begin{equation}\label{constraint3}
\begin{aligned}
    \text{s.t. } & D_H^w(H_i, H_j) - D_H^w(H_i, H_k) \geq 1-\xi_{ijk}, \\
              & \forall(i,j,k) \; c_i \neq c_j \: \wedge \: c_i = c_k
\end{aligned}
\end{equation}

\begin{equation}\label{constraint4}
\begin{aligned}
     &  |D_H^w(H_{i'}, H_{j'}) - D_H^w(H_{i'}, H_{k'})| \leq \gamma_{{i'}{j'}{k'}}, \\
     &  \forall({i'},{j'},{k'}) \; c_{i'} = c_{j'} = c_{k'}
\end{aligned}
\end{equation}

\begin{equation}\label{constraint5}
    \xi_{ijk} \geq 0; \gamma_{{i'}{j'}{k'}} \geq 0; w \geq 0
\end{equation}
where (\ref{constraint3}) indicates inter-class distance should be big enough and (\ref{constraint4}) prefers small intra-class distance.
C is the trade-off parameter between maximizing the margin and satisfying the distance preference. The weights $w$ used in the weighted Hamming distance for binary codes nicely fit into the SVM weights. Note the ranking here is essentially used to classify a sample pair into semantically similar or not similar. But this formulation is general enough that finer ranking between samples from the same class could also be imposed to give more detailed instance level ranking. The way $w$ works is very flexible. Supervision could be encoded in the sense of same object category or by user feedback, e.g. clicked image search results to give finer ranking within a specific object category.

We could reformulate the problem to move the ranking constraints into cost terms in the objective function.

\begin{equation}\label{opt2}
\begin{aligned}[l]
    \min & \frac{1}{2}\|w\|_2^2 + \\
         & C_{\xi} \sum \max\{0, 1-[D_H^w(H_i, H_j) - D_H^w(H_i, H_k)]\}^2 + \\
         & C_{\gamma} \sum[D_H^w(H_{i'}, H_{j'}) - D_H^w(H_{i'}, H_{k'})]^2 \\
         & \;\;\;\;\;\;\; \forall(i,j,k)\; c_i \neq c_j \: \wedge \: c_i = c_k \\
         & \;\;\;\;\;\;\; \forall({i'},{j'},{k'})\; c_{i'} = c_{j'} = c_{k'} \\
         & \;\;\;\;\;\;\; w \geq 0
\end{aligned}
\end{equation}


Here we restrict weights $w$ to be non-negative values. Although ranking itself doesn't impose such constraint, this is still necessary to ensure a valid non-negative distance measure.

By penalizing the squared loss for slack variables and converting it to an unconstraint form, this optimization problem becomes a Quadratic Optimization (QP) with non-negativity constraint. There are multiple ways to solve it. Gradient Descent is the simplest method and it has been used to solve this type of problems in ~\cite{largescaleranking} and ~\cite{bottou2010large}. Newton's method was also applied to efficiently solve it in~\cite{primalsvm} and~\cite{relataive}. In our case, to force $w$ to be non-negative, Exponentiated Gradient Descent (EGD) and Projected Gradient Descent (PGD) are both able to optimize it. EG updates the weights by multiplying a exponential function of the gradient and PGD only do update if the updated value is still non-negative. Both methods can ensure $w$ to be non-negative after every update. We found in the experiments EGD generally gave better performance compared with PGD.

Assume the cost function is denoted as $J(w)$, at every iteration, after computing the gradient $\frac{\partial J}{\partial w_i}$, the $i$th component of $w$ is updated as:

\begin{equation}\label{eg}
\begin{aligned}[l]
    & w_i = w_i*e^{-\eta \frac{\partial J}{\partial w_i}}
\end{aligned}
\end{equation}
where $\eta$ is a tunable parameter working as a learning rate.

Exponentiated Gradient Descent is an example of multiplicative update, which typically leads to faster convergence than additive updates and naturally produce non-negative solution. Besides, sparse solution could be achieved using EGD, which reveals least important binary code bits in our problem.

\subsubsection{Extension to Online Updating}

Online learning has the benefit of incrementally incorporating information from newly observed data so that the model learned before could be improved over time and scale to larger data set. This opens the door to real world applications. For example, in image search, after getting a list of candidate images, user will click the ones he/she likes. This implicitly poses a preference ranking for the input query. The ability to learn from this useful feedback helps to make the future search results closer to user's expectation.

There are relatively few work about directly generating binary code on-the-fly, one recent example is~\cite{onlinehash}. Although the hash function is able to update incrementally, it is still unavoidable to recompute binary codes for all data set samples. On the contrary, in our case, we shift the update target from binary codes to weights. The code base is not required to change, only a few weight values need to be updated. Online training on SVM-like maximum margin learners allow training as a step-by-step process by giving a single sample at a time and shows promising accuracy compared to standard batch training mode while largely boosting the computational efficiency. This advantage not only enables training on new data but also makes the learning scalable for very large data set since the data set is not required to store in the memory during the training process. Among these methods, Stochastic Gradient Descent~\cite{bottou2010large} and Passive-Aggressive algorithm~\cite{crammer2006online} are the most popular ones.

We use the Passive-Aggressive paradigm to formulate the online version of the weight learning. Assume at time t, the weight vector is $w(t)$, and given a new set of rank pairs with one similar pair and one dissimilar pair, we solve for a new $w$ to satisfy the new ranking constraints.
The whole problem is as follows:
\begin{equation}\label{online_opt}
\begin{aligned}[l]
    \min & \frac{1}{2}\|w-w(t)\|_2^2 + \\
         & C_{\xi} \: max\{0, 1-[D_H^w(H_i, H_j) - D_H^w(H_i, H_k)]\}^2 + \\
    	 & C_{\gamma} \: [D_H^w(H_{i'}, H_{j'}) - D_H^w(H_{i'}, H_{k'})]^2 \\
         & c_i \neq c_j, c_i = c_k \\
         & c_{i'} = c_{j'} = c_{k'} \\
         & w \geq 0
\end{aligned}
\end{equation}

%
%
The non-negative constraint is also satisfied by using Exponentiated Gradient Descent. We give the algorithm diagram for online weight learning at Algorithm 1.

\begin{figure*}[ht!]
\centering
\includegraphics[width=\linewidth]{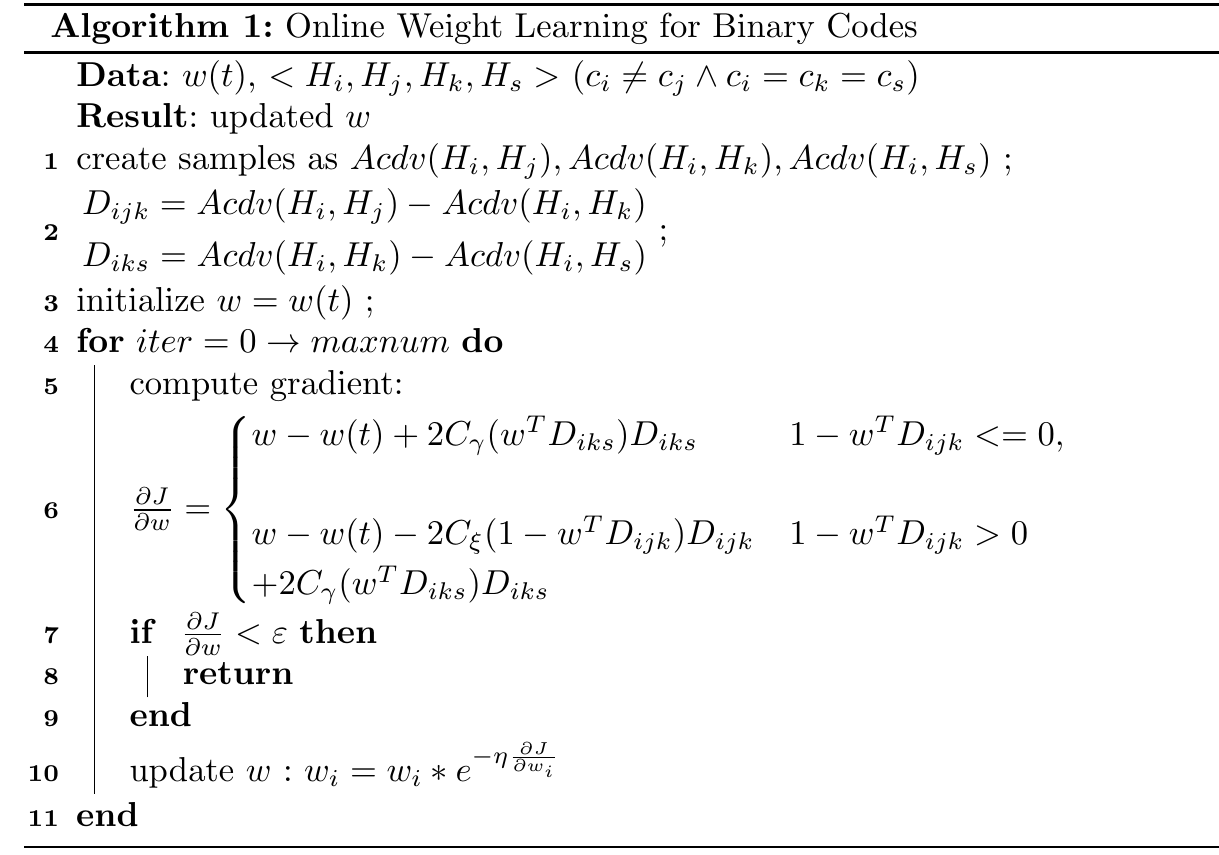}
\label{online_algo}
\end{figure*}

%

Practically, to ensure a more reliable gradient estimate, instead of using a single sample at a time to update $w$, we use mini-batch EGD to consider more samples while maintaining the efficiency of updating at each iteration. In our experiment, the mini-batch is selected as 10 samples and the update usually takes only 2-3 iterations.

\subsection{Training Data Generation}
\label{train_data}

One problem in weight learning using ranking data is that a huge number of similar or dissimilar pairs could be generated given a training set. This especially come into consideration when we are using triplets for dissimilarity and similarity constraints. Since our goal is to learn query independent weights for ranking, to ensure a full coverage of the data set across different classes, random sampling under Multinomial distribution is applied to extract a given number of training samples. The class-sample ratio in the whole data set is treated as event probability of the distribution. Assume a data set is given with pre-computed binary code for each sample in each class, $I=\{\langle H_i, c_i \rangle\}^N_{i=1}$, $H_i$ is the binary code for the $i$th image feature $x_i$, $c_i$ is the corresponding class label. First, a query point $q_i$ is sampled from one class, then two similar points $q_k$ and $q_s$ are selected from the same class and a dissimilar one $q_j$ is also uniformly sampled from a different class, resulting a quadruplet $(q_i, q_j, q_k, q_s)$ where $c_i \neq c_j, c_i = c_k = c_s$. Then this quadruplet is converted to ACDVs as used in our Ranking SVM optimization: $\langle Acdv(q_i, q_j), Acdv(q_i, q_k), Acdv(q_i, q_s) \rangle$. Of course, we can further extract more ACDVs by taking difference between other combinations of sample binary codes. This ACDV triplet offers balance for similar and dissimilar pairs and is sufficient for learning the weights. The process is done with non-replacement of samples to ensure no duplicated training samples exist. Finally, we have a collections of training triplets: $\{\langle Acdv(q_i, q_j), Acdv(q_i, q_k), Acdv(q_i, q_s)\rangle | c_i \neq c_j, c_i = c_k = c_s \}$.

\section{Experiments}

\textbf{Data Sets:} To demonstrate the benefit of our bitwise weight learning algorithm, we use three image data sets for evaluation.
The first one is MNIST, a hand-written digit database collected in~\cite{minst}. This data set contains 60000 training samples and 10000 test samples for ten digits (0 - 9). Each sample is a 28X28 grayscale image. The second data set is the CIFAR10 data set~\cite{cifat10}, which contains 60000 real world color images in 10 classes, with 6000 images per class. These classes include objects like airplane, bird, dog, ship etc. Each image is resized to 32x32. The last one is YouTube Faces Database~\cite{youtubeface}. This data set has 3425 videos of 1595 people. Each video has averagely 181.3 frames. The face in each frame is annotated by a bounding box. All the face images for a person is treated as a class.

\textbf{Binary Code Bases:} We select several hashing methods to produce base binary codes on which our weight learning is applied. These methods include Locality Sensitive Hashing (LSH), Spectral Hashing (SH), Isotropic Hashing (ISOH) and Iterative Quantization (ITQ). The implementations of these methods are provided by the authors. For MNIST and YouTube Faces, the raw feature is just the vectorized pixel values from each grayscale digit image. For CIFAR10, we use GIST~\cite{gist} to compute a global image feature as in~\cite{smallcodes}. Selected hashing methods use these raw features to produce individual set of binary codes on each data set. Codes of 16, 32, 64, 98, 128 bits are tested.

\textbf{Evaluation:} Since the weighting is eventually applied with Hamming distance to produce a real-valued score for ranking, we evaluate the algorithm as in image retrieval problems. Each data set is divided into training set and test set evenly. The training set is used to learn the hash functions for producing binary code bases. Also, ranking pairs are sampled from the training set to learn the weighting vectors for each type of code with different lengths. A set of queries are randomly selected from each class in the test set. For a given query $q$, we compute its binary code, and use the learned bitwise weights to measure distance with all samples in the test set, then these samples are ranked in ascending order to form a list. Note that when used in practical applications, this weighted Hamming distance can combine with standard Hamming distance in hash table probing or binary code scanning using XOR operation, thus it only needs to perform on small number of fetched neighbor candidates and is still very efficient to compute.
At each list location $i$, we compute its precision (P) and recall (R) values as:
\begin{equation}\label{precision}
\begin{split}
    Precision@i = \frac{|\{x_j|c_{x_j} = c_q, j<i\}}{i}
\end{split}
\end{equation}
\begin{equation}\label{recall}
\begin{split}
    Recall@i = \frac{|\{x_j|c_{x_j} = c_q, j<i\}}{|\{x|c_x=c_q\}|}
\end{split}
\end{equation}

To aggregate ranking results for all the query points, we use average PR curve by computing the average of all PR values at each location. We also calculate the Mean Average Precision (MAP) value for each combination of code type and code length, following the PASCAL VOC style for computing average precision (AP)~\cite{pascal}.
Throughout all experiments, our method is denoted with a postfix 'weighted' to indicate a weighted version of the Hamming distance for binary code ranking.

To show the advantage of our weight learning algorithm, we compare it with all standard binary code using Hamming distance. We further compare with the recently proposed WhRank which also apply weighting in binary code distance measure. The WhRank is implemented using the simpler version (WhRank1) which is claimed in~\cite{weightedhamming} to achieve similar performance with the full WhRank.

\subsection{Experimental Results}

\begin{figure*}[ht!]
\centering
\includegraphics[width=\linewidth]{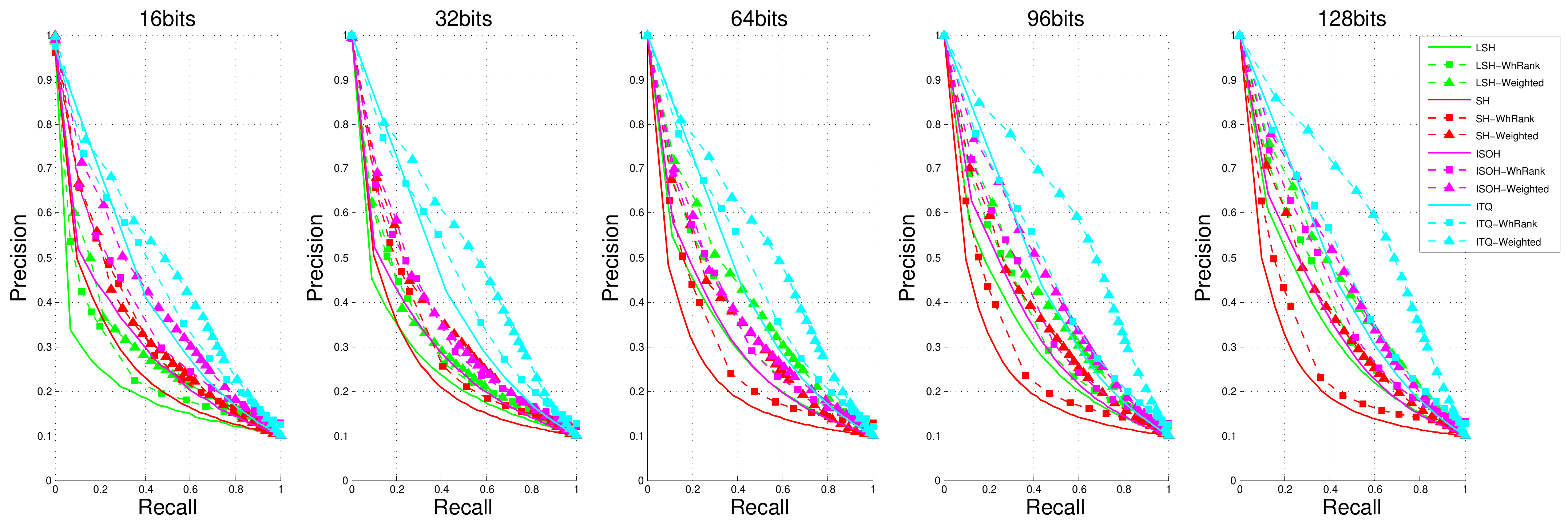}
\caption{(Best viewed in color) Precision-Recall curves on MNIST data set.}
\label{mnist_res}
\end{figure*}

\textbf{MNIST:}\label{mnist_exp}
We randomly select 10\% samples from each digit class in the test set to construct the query set. The ranking set is generated as Sec~\ref{train_data} and 5000 triplets are selected from the training images to learn the weights in the offline mode. The average Precision-Recall curves for each binary code type are depicted in Fig~\ref{mnist_res}. Our learned weights consistently improve the ranking performance for each type of binary code, average performance gain ranges from 5\% to 20\%.
The weak hashing methods like LSH and SH could be boosted to compete with better hashing methods like ITQ and ISOH. ITQ achieves surprisingly good results with bitwise weighting, which is likely because ITQ produces high quality base codes on MNIST which allows the weight learning to better separate similar codes and dissimilar ones.
MAPs of all code combination is shown in Table~\ref{map_mnist}. From the table, we could see most weighted versions get better performance with the increasing length of code bits. Weighted ITQ tops in all combinations of bit lengths.
WhRank is able to bring improvement to base codes while our method outperforms WhRank in a relatively large margin.

\begin{table}[t]
\centering
\begin{tabular}{cccccc} \hline
        \textbf{Method} & \textbf{16bit} & \textbf{32bit} & \textbf{64bit} & \textbf{96bit} & \textbf{128bit} \\ \hline
                    LSH & $0.233$ & $0.272$ & $0.312$ & $0.321$ & $0.342$ \\
             LSH-WhRank & $0.264$ & $0.309$ & $0.353$ & $0.361$ & $0.388$ \\
           LSH-Weighted & $0.306$ & $0.311$ & $0.410$ & $0.404$ & $0.432$ \\
                     SH & $0.268$ & $0.271$ & $0.246$ & $0.250$ & $0.248$ \\
              SH-WhRank & $0.335$ & $0.321$ & $0.287$ & $0.286$ & $0.282$ \\
            SH-Weighted & $0.340$ & $0.354$ & $0.351$ & $0.376$ & $0.375$ \\
                   ISOH & $0.309$ & $0.307$ & $0.319$ & $0.347$ & $0.355$ \\
            ISOH-WhRank & $0.353$ & $0.350$ & $0.361$ & $0.385$ & $0.399$ \\
          ISOH-Weighted & $0.402$ & $0.368$ & $0.377$ & $0.437$ & $0.447$ \\
                    ITQ & $0.326$ & $0.336$ & $0.337$ & $0.342$ & $0.354$ \\
             ITQ-WhRank & $0.430$ & $0.441$ & $0.447$ & $0.446$ & $0.452$ \\
           ITQ-Weighted & $\mathbf{0.466}$ & $\mathbf{0.497}$ & $\mathbf{0.502}$ & $\mathbf{0.561}$ & $\mathbf{0.577}$ \\ \hline
\end{tabular}
\caption{MAPs for different methods on MNIST data set.}
\label{map_mnist}
\end{table}

\begin{figure*}[t]
\centering
\includegraphics[width=\linewidth]{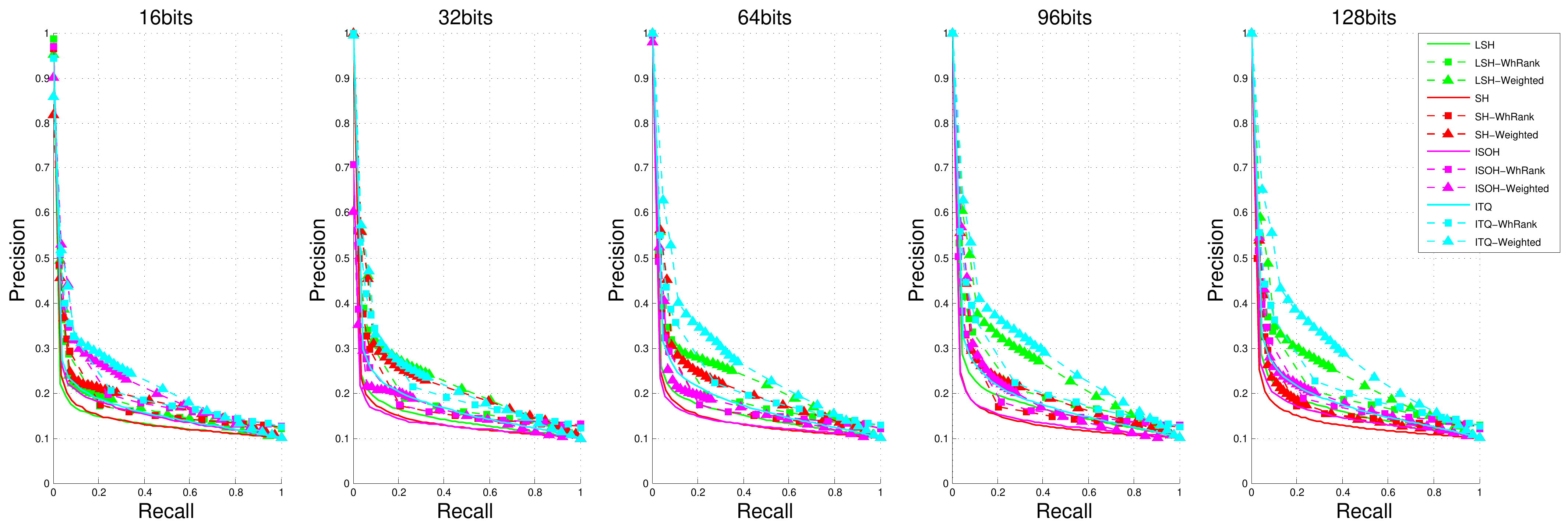}
\caption{(Best viewed in color) Precision-Recall curves on CIFAR10 data set.}
\label{cifar_res}
\end{figure*}

\textbf{CIFAR10:}
The CIFAR10 data set is quite different from MNIST, which has relatively large variations within the same class, making it a lot more difficult. We follow the same testing steps in MNIST by picking 10\% samples from each class in test set as query set, and use 8000 triplets from training set for weight learning. PR curves are drawn in Fig.~\ref{cifar_res} and MAPs are shown in Table~\ref{map_cifar}. We are still able to get about $5\% \sim 10\%$ average precision improvement under the same recall value in this hard data set. The weight learning performs fairly stable across different base code types and code lengths. Interesting observations are noticed from this graph. The performance gain for each binary code type varies differently. The simplest LSH is able to show quite significant boost with our learned weights compared with other binary code types. This might indicate that for data set with high complexity and class variance, randomly generated binary codes are relatively more flexible to embed supervision since it doesn't enforce strong relations between the raw features and corresponding codes.

\begin{table}[t]
\centering
\begin{tabular}{cccccc} \hline
        \textbf{Method} & \textbf{16bit} & \textbf{32bit} & \textbf{64bit} & \textbf{96bit} & \textbf{128bit} \\ \hline
                    LSH & $0.195$ & $0.203$ & $0.209$ & $0.220$ & $0.215$ \\
             LSH-WhRank & $0.215$ & $0.225$ & $0.231$ & $0.241$ & $0.239$ \\
           LSH-Weighted & $0.215$ & $0.261$ & $0.265$ & $0.284$ & $0.272$ \\
                     SH & $0.193$ & $0.195$ & $0.196$ & $0.195$ & $0.195$ \\
              SH-WhRank & $0.212$ & $0.216$ & $0.215$ & $0.214$ & $0.213$ \\
            SH-Weighted & $0.216$ & $0.243$ & $0.247$ & $0.240$ & $0.212$ \\
                   ISOH & $0.207$ & $0.166$ & $0.196$ & $0.198$ & $0.207$ \\
            ISOH-WhRank & $0.229$ & $0.189$ & $0.216$ & $0.219$ & $0.226$ \\
          ISOH-Weighted & $0.242$ & $0.180$ & $0.216$ & $0.225$ & $0.230$ \\
                    ITQ & $0.209$ & $0.222$ & $0.226$ & $0.227$ & $0.227$ \\
             ITQ-WhRank & $0.231$ & $0.242$ & $0.254$ & $0.255$ & $0.258$ \\
           ITQ-Weighted & $\mathbf{0.248}$ & $\mathbf{0.265}$ & $\mathbf{0.291}$ & $\mathbf{0.305}$ & $\mathbf{0.305}$ \\ \hline
\end{tabular}
\caption{MAPs for different methods on CIFAR10 data set.}
\label{map_cifar}
\end{table}

\begin{figure*}[t]
\centering
\includegraphics[width=\linewidth]{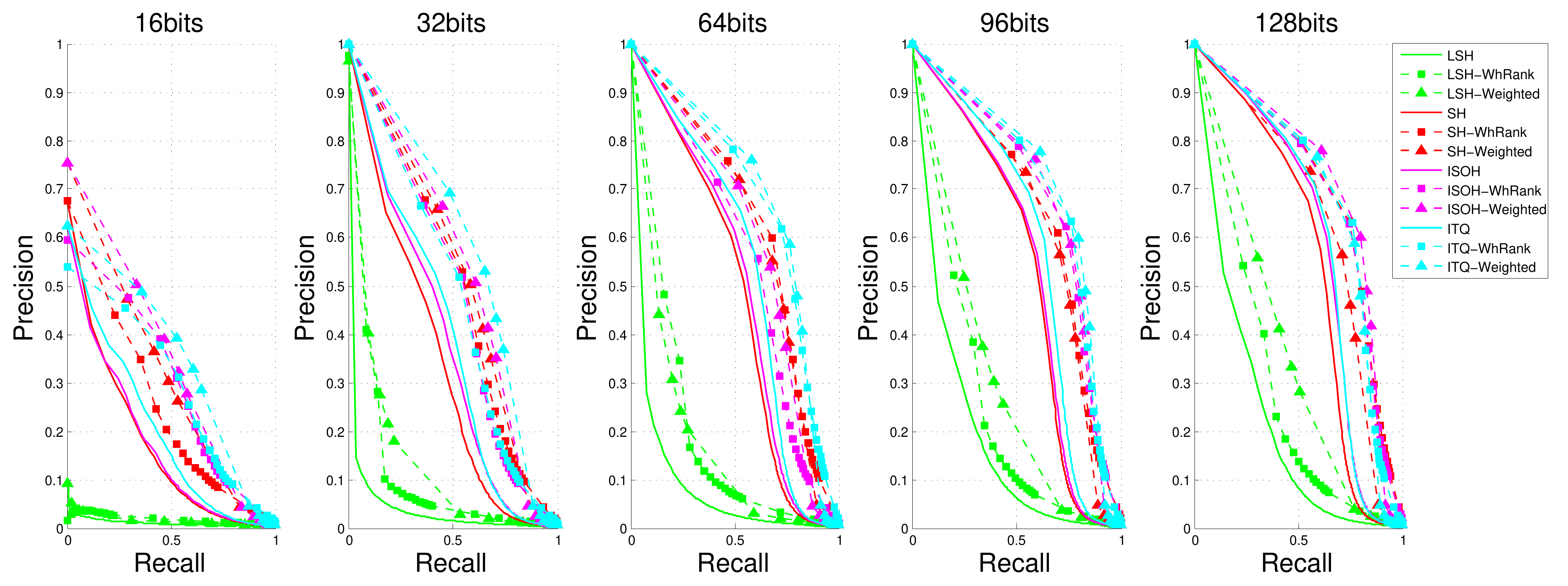}
\caption{(Best viewed in color) Precision-Recall curves on YouTube Faces data set. The Weighted suffix means the ranking performance with weighted Hamming distance.}
\label{face_res}
\end{figure*}

\textbf{YouTube Faces:}
This data set has much larger scale compared to the previous two. To make the testing process more efficient, we pick 5\% samples from each test class to form query set. 10000 triplets are used to learn bit weights. The PR-curves are shown in Fig~\ref{face_res} and Table~\ref{map_face}. Due to the nice annotation of face patches in each image and the fact that these images are adjacent in the original video clip, making them very similar to each other. Performance of several methods in this data set are satisfying. Weighted ISOH and weighted ITQ reach similar performance with the code length increases. WhRank combines ITQ outperforms other combination with 96 bit codes and weighted ISOH achieves best performance with 128 bit. This insight may lead to the question of how well the supervision information could be added to binary codes generated with optimizing different objectives. This experiment demonstrates our weight learning method is able to help even in the case of a large scale data set.

\begin{table}
\centering
\begin{tabular}{cccccc} \hline
        \textbf{Method} & \textbf{16bit} & \textbf{32bit} & \textbf{64bit} & \textbf{96bit} & \textbf{128bit} \\ \hline
                    LSH & $0.017$ & $0.108$ & $0.136$ & $0.200$ & $0.232$ \\
             LSH-WhRank & $0.019$ & $0.142$ & $0.204$ & $0.240$ & $0.273$ \\
           LSH-Weighted & $0.018$ & $0.147$ & $0.178$ & $0.266$ & $0.327$ \\
                     SH & $0.180$ & $0.320$ & $0.450$ & $0.508$ & $0.531$ \\
              SH-WhRank & $0.239$ & $0.444$ & $0.551$ & $0.568$ & $0.597$ \\
            SH-Weighted & $0.275$ & $0.447$ & $0.530$ & $0.552$ & $0.553$ \\
                   ISOH & $0.178$ & $0.361$ & $0.464$ & $0.515$ & $0.557$ \\
            ISOH-WhRank & $0.282$ & $0.430$ & $0.499$ & $0.609$ & $0.625$ \\
          ISOH-Weighted & $0.307$ & $0.468$ & $0.516$ & $0.594$ & $\mathbf{0.627}$ \\
                    ITQ & $0.204$ & $0.377$ & $0.510$ & $0.550$ & $0.563$ \\
             ITQ-WhRank & $0.273$ & $0.432$ & $0.592$ & $\mathbf{0.627}$ & $0.613$ \\
           ITQ-Weighted & $\mathbf{0.294}$ & $\mathbf{0.496}$ & $\mathbf{0.593}$ & $0.623$ & $0.605$ \\ \hline
\end{tabular}
\caption{MAPs for different methods on YouTube Faces data set.}
\label{map_face}
\end{table}

\begin{figure}[t]
\subfigure[MNIST]{
\includegraphics[width=0.3\textwidth, trim=0 0 5 0]{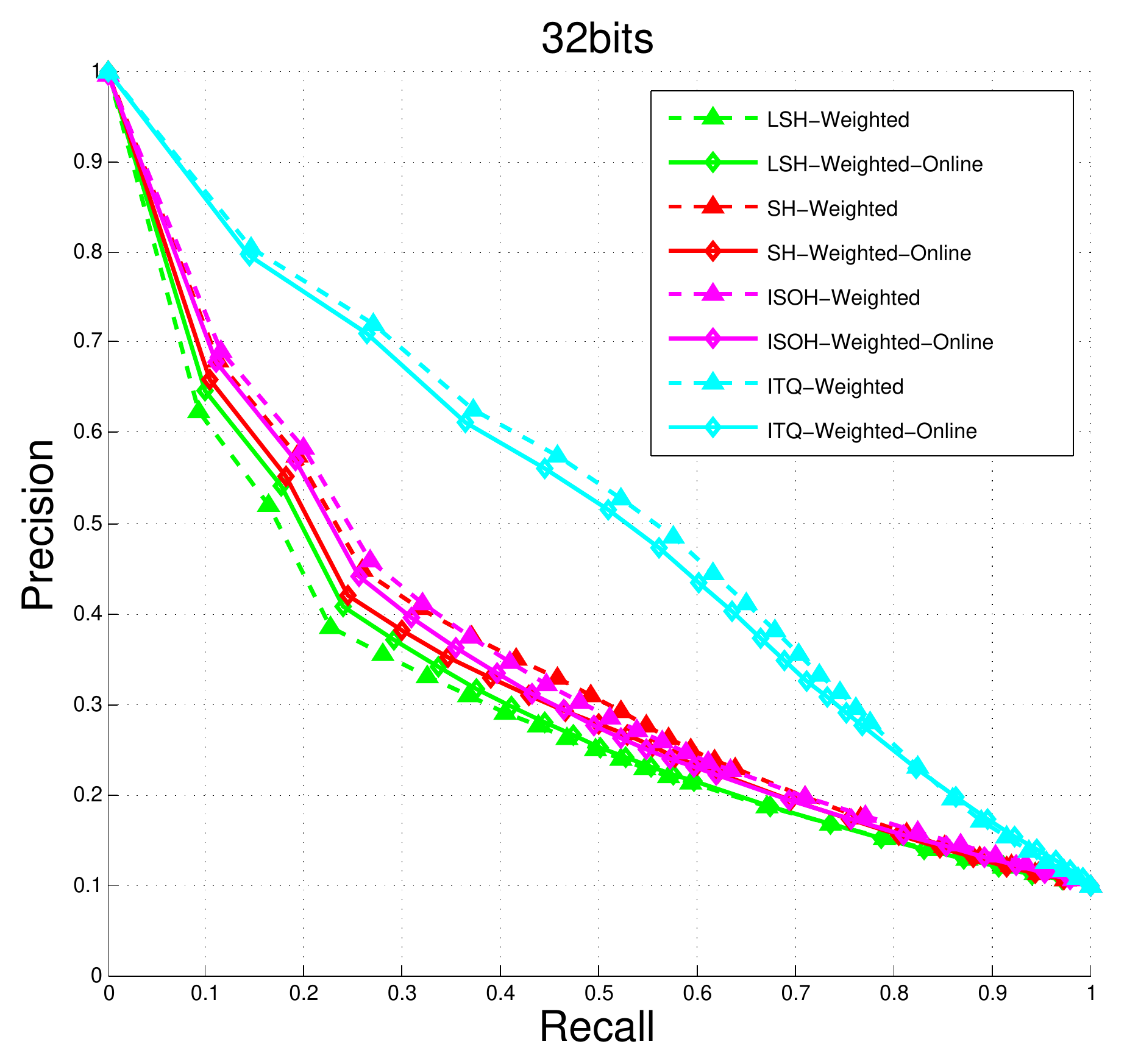}
\label{mnist_online}
}
\subfigure[CIFAR10]{
\includegraphics[width=0.3\textwidth, trim=0 0 5 0]{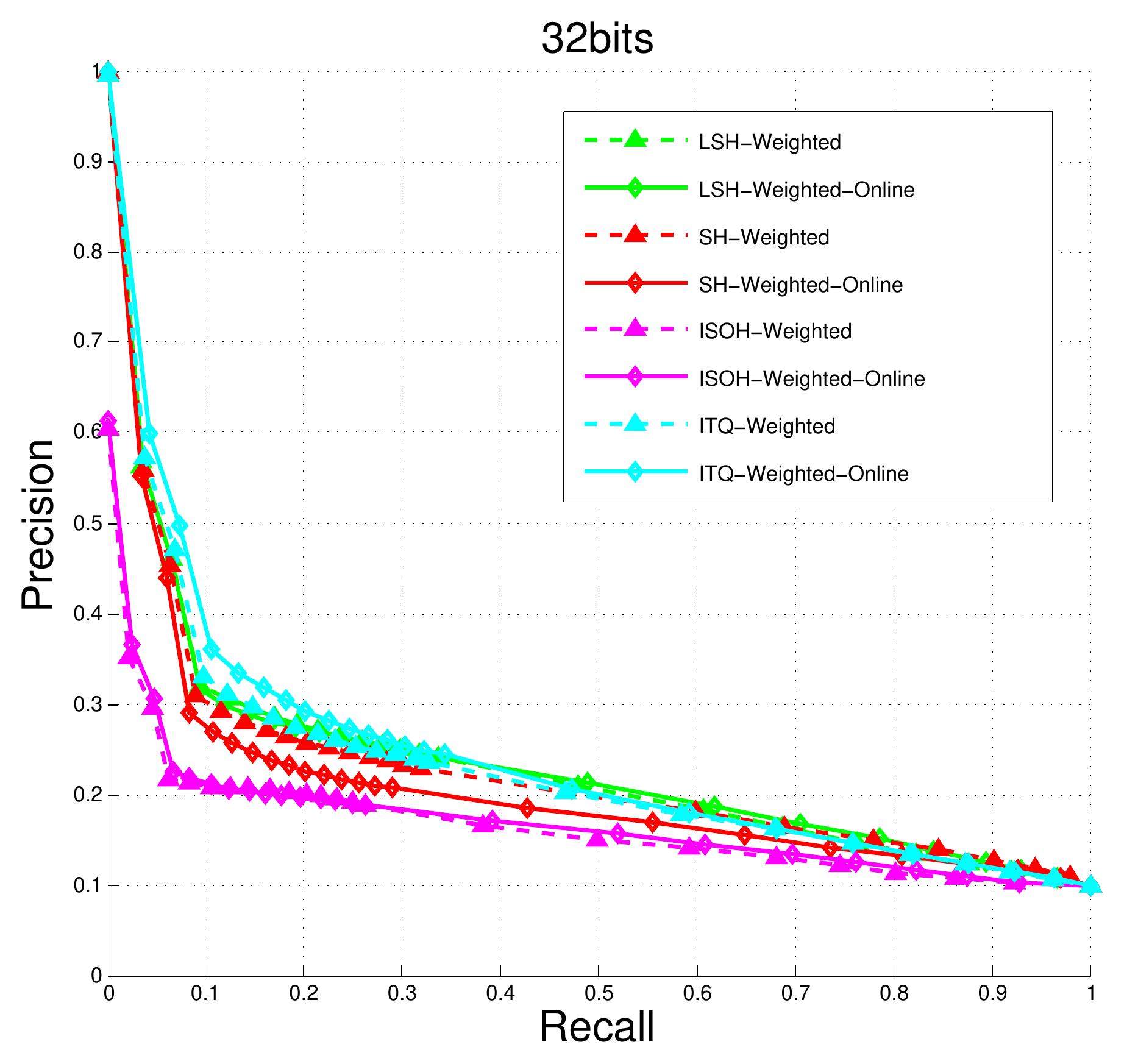}
\label{cifar_online}
}
\subfigure[YouTube Faces]{
\includegraphics[width=0.3\textwidth, trim=0 0 5 0]{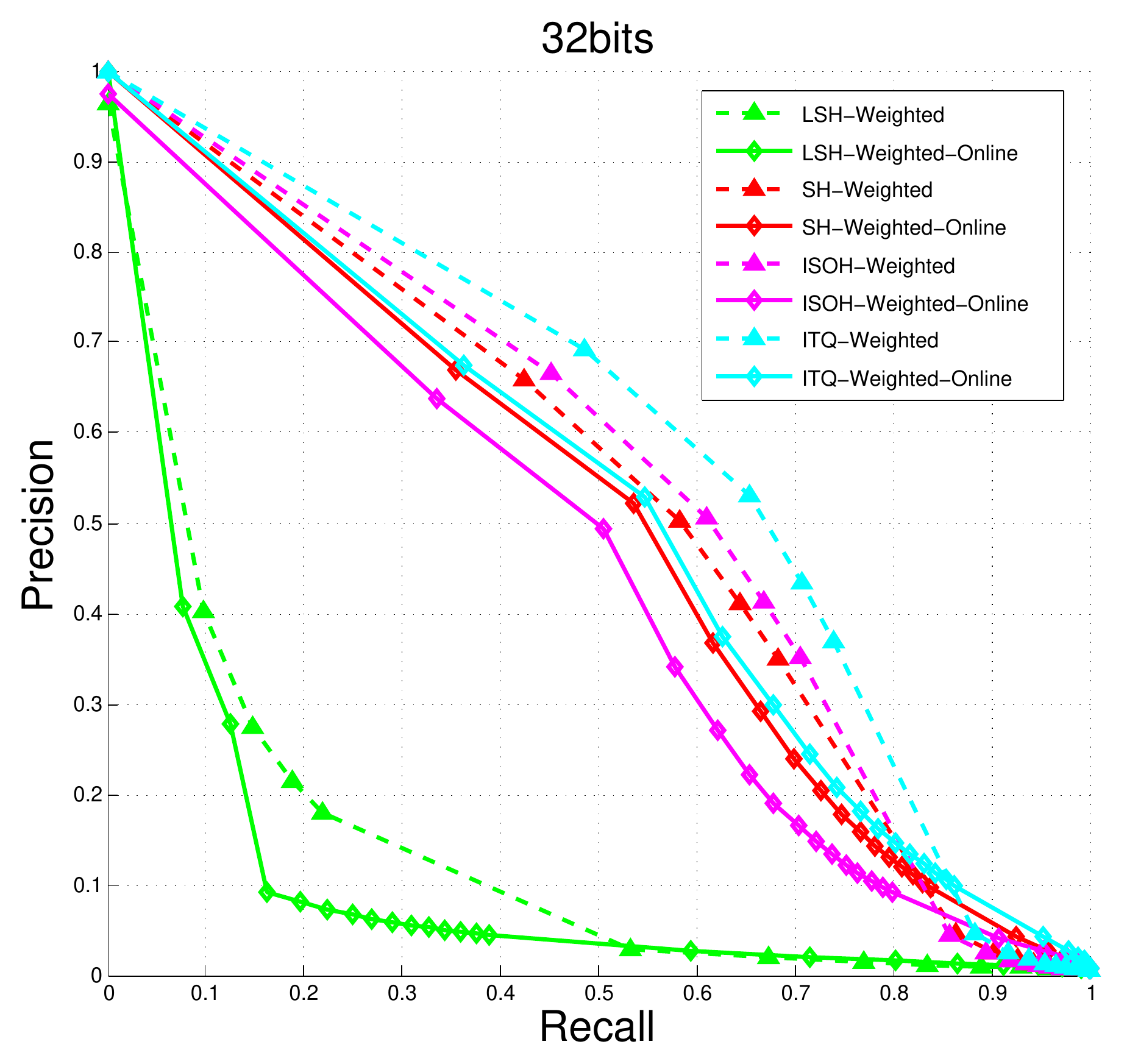}
\label{face_online}
}
\caption{Comparison between offline and online learned weights in MNIST, CIFAR10, YouTube Faces data sets with 32 bits code.}
\label{online_weights}
\end{figure}

\textbf{Offline Mode vs. Online Mode:}
To demonstrate the effectiveness of online weight learning using the Ranking SVM model, we conducted comparison between offline learned and online learned weights on ranking the query samples on all data sets. Since the main purpose is to investigate the difference between these two weight learning methods, we ignore unweighted baseline performance which have been shown previously. 32bit code is used to perform the comparison.
In this experiment, the online training process uses the same amount of training data as offline training with the exception that the data is fed to the algorithm one by one (or a small set) at a time. The results are shown in Fig~\ref{online_weights}.

 From the graph, it shows very similar performance between offline and online learned weights on MNIST and CIFAR10. On YouTube Faces data set, it drops about 8\% for ITQ and ISOH. In general, online weight updating gives reasonable performance compared with its offline counterpart which makes it a good alternative for this weight learning algorithm in a more scalable and flexible way. Extra benefits of online method are faster iteration and much less memory. We are now not required to store all the training data at one place and run heavy iterations on all of them. The updating process removes the computational burden while adapt the new weights to new input data.

\section{Conclusions}

In this paper, we proposed a bitwise weight learning method over binary hash codes
given supervised rank information. This post-hashing procedure successfully integrates
the supervised information into a pre-defined binary-code base and thus endows binary codes
with the discriminative power. Extensive experiments have demonstrated that
our method can accommodate for different types of binary codes and achieve satisfying ranking performance.
Consequently, our method provides a promising way to enhance the Hamming distance measure
so that the ranking of the search results appears more semantically meaningful.
The online learning scheme further increases the flexibility of our method,
permitting rapid incremental updates for newly coming data.
Such an advantage would make the proposed method apply to more real-world applications like interactive image search.

\clearpage

\bibliographystyle{splncs03}
\bibliography{egbib}
\end{document}